\documentclass{article} 
\usepackage{nips15submit_e,times}
\usepackage[bookmarks=false]{hyperref}
\usepackage{url}
\usepackage{graphicx} 

\usepackage{amsmath}
\usepackage{amsfonts}
\usepackage{amsthm}
\usepackage{caption}
\usepackage{subcaption}

\usepackage{dsfont}

\usepackage{caption}
\DeclareCaptionType{copyrightbox}
\usepackage{subcaption}

\newcommand{\ones}{\ensuremath{\mathds{1}}}
\newcommand{\R}{\ensuremath{\mathds{R}}}

\newcommand{\va}{\boldsymbol{\alpha}}

\newcommand{\Bx}{\mathbf{x}}

\renewcommand{\vec}[1]{\mathbf{#1}}

\usepackage{algorithm}
\usepackage{algorithmic}

\title{Doubly stochastic large scale kernel learning \\ with the empirical kernel map}

\author{
Nikolaas Steenbergen\\
DFKI, Berlin, Germany\\
\texttt{nikolaas.steenbergen@dfki.de}\\
\And
Sebastian Schelter\\
TU Berlin, Berlin, Germany\\
\texttt{sebastian.schelter@tu-berlin.de}\\
\And
Felix Biessmann
TU Berlin, Berlin, Germany\\
\texttt{felix.biessmann@tu-berlin}\\
}

\nipsfinalcopy 

\begin{document}

\maketitle

\begin{abstract} 
With the rise of big data sets, the popularity of kernel methods declined and neural networks took over again. The main problem with kernel methods is that the kernel matrix grows quadratically with the number of data points. Most attempts to scale up kernel methods solve this problem by discarding data points or basis functions of some approximation of the kernel map. Here we present a simple yet effective alternative for scaling up kernel methods that takes into account the entire data set via doubly stochastic optimization of the emprical kernel map. The algorithm is straightforward to implement, in particular in parallel execution settings; it leverages the full power and versatility of classical kernel functions without the need to explicitly formulate a kernel map approximation. We provide empirical evidence that the algorithm works on large data sets. 
\end{abstract}

\section{Introduction\vspace{-0.1in}}
\indent When kernel methods \cite{Muller:2001p2592,shawe2004kernel} were introduced in the machine learning community, they quickly gained popularity and became the gold standard in many applications. A reason for this was that kernel methods are powerful tools for modelling nonlinear dependencies. Even more importantly, kernel methods offer a convenient split between modelling the data with an expressive set of versatile kernel functions for all kinds of data types (e.g.,~graph data~\cite{shawe2004kernel} or text data~\cite{John2000}), and the learning part, including both the learning paradigm (unsupervised vs. supervised) and the optimization. 

The main drawback of kernel methods is that they require the computation of the kernel matrix $\vec{K}\in\R^{N\times N}$, where $N$ is the number of samples. For large data sets this kernel matrix can neither be computed nor stored in memory. Even worse, the learning part of kernel machines often has complexity $\mathcal{O}(N^3)$. This renders standard formulations of kernel methods intractable for large data sets. When machine learning entered the era of web-scale data sets, artificial neural networks, enjoying learning complexities of $\mathcal{O}(N)$, took over again, and have been dominating the top ranks of competitions, the press on machine learning and all major conferences since then. But the advantage of neural networks -- or other nonlinear supervised algorithms that perform well on large data sets in many applications, such as Random Forests~\cite{Breiman2001} -- leaves many researchers with one question (see e.g.,~\cite{Lu2014}): What if kernel methods could be trained on the same amounts of data that neural networks can be trained on? 

There have been several attempts to scale up kernel machines, most of which fall into two main categories: a) approximations of the kernel map based on subsampling of Fourier basis functions (see~\cite{Rahimi2008}) or b) approximations of the kernel matrix based on subsampling data points (see~\cite{Williams2000}).
While both of these are powerful methods which often achieve competitive performance, most applications of these approximations solve the problem of scaling up kernel machines by discarding data points or Fourier basis functions from the computationally expensive part of the learning. We present a remarkably simple yet effective alternative of scaling up kernel methods that -- in contrast to many previous approaches -- allows us to make use of the entire data set. 

Similar to~\cite{Dai2014} we propose a doubly stochastic approximation to scale up kernel methods. In contrast to their work however, who use an {\em explicit} approximation of the kernel map\footnote{We follow the convention that {\em explicit} kernel maps refer to a data independent kernel map approximation, see also \autoref{sec:kernels}.}, we propose to use an approximation of the {\em empirical} kernel map. While the optimization follows a similar schema, there is evidence suggesting that approximations of the explicit kernel map can result in lower performance~\cite{Yang2012}.
The approach is called doubly stochastic because there are two sources of noise in the optimization: a) the first source samples random data points at which a noisy gradient of the dual coefficients is evaluated and b) the second source samples data points at which a noisy version of the empirical kernel map is evaluated. We propose a redundant data distribution scheme that allows for computing approximations that go beyond the block-diagonal of the full kernel matrix, as proposed in~\cite{Deisenroth2015} for example. We perform experiments on synthetic data comparing the proposed approach with other approximations of the kernel map, and conduct experiments with a parallel implementation to show the scale-up behaviour on a large data set. 

In the following, we give a short summary of the essentials on kernel machines, in \autoref{sec:large_scale_kernel_learning} we give a broad overview over other attempts to scale up kernel methods, \autoref{sec:dskl} outlines the main idea of the paper and \autoref{sec:experiments} describes our experiments.

\section{Kernel methods}\label{sec:kernels}
This section summarizes some of the essentials of kernel machines. For the sake of presentation we only consider supervised learning and assume $D$-dimensional real-valued input data $\Bx_i\in\R^D$ and a corresponding boolean label $y_i\in\{-1,1\}$. The key idea of kernel methods is that the function to be learned $\phi^*(\Bx_i) $, evaluated at the $i$th data point $\Bx_i$, is modelled as a linear combination $\va\in\R^N$ of similarities between data point $\vec{x}_i$ and data points $\vec{x}_j,~j=\{1,2,\dots,N\}$, both mapped to a potentially infinite dimensional {\em kernel feature space} $\mathcal{S}$

\begin{align}\label{eq:kernel_trick}
\phi^*(\Bx_i)=\sum_{j=1}^N k(\Bx_i,\Bx_j)\alpha_j.
\end{align}
Here $N$ again denotes the number of data points in the data set, $\alpha_i$ denotes the $i$-th entry of $\va$, and the kernel function $k(.,.)$ measures the similarity of data points in the kernel feature space by computing inner products in $\mathcal{S}$
\begin{align}\label{eq:kernel_function}
k(\Bx_i,\Bx_j)=\langle \phi(\Bx_i), \phi(\Bx_j)\rangle_{\mathcal{S}}.
\end{align}
Kernel methods became popular as a nonlinear dependency between data points (and labels) in input space becomes linear in $\mathcal{S}$.
Taking the shortcut through $k(.,.)$, i.e., mapping data points to $\mathcal{S}$ and computing their inner products without ever formulating the mapping $\phi$ {\em explicitely} when learning $\phi$ is sometimes referred to as {\em kernel trick}. Methods that attempt to construct an {\em explicit} representation of $\phi$ are hence sometimes referred to as {\em explicit kernel approximations}~\cite{Dai2014}. 

Most kernel machines then minimize a function $\mathcal{E}(y, \Bx, \va, k)$ which combines a loss function $l(y, \Bx, \va, k)$ with a regularization term $r(\va)$ that controls the complexity of $\phi$ 
\begin{align}\label{eq:error_function}
\mathcal{E}(y, \Bx, \va, k) = r(\va) + l(y, \Bx, \va, k).
\end{align}
The regularizer $r(\va)$ often takes the form of some $\mathcal{L}_p$ norm of the vector of dual coefficients $\va$, where usually $p=2$. A popular example of \autoref{eq:error_function} is that of the kernel support-vector machine (SVM) \cite{Cortes1995}: a hinge loss combined with a quadratic regularizer
\begin{equation}\label{eq:svm_gradient}
\mathcal{E}_{SVM} = ||\max \left(0,\ones-\text{diag}(\vec{y}) \vec{K} \va \right)||_1 + \lambda ||\va||_2, \quad \frac{\partial\mathcal{E}_{SVM}}{\partial \va} =\max \left(0,1-\left(\lambda\va - \vec{y}^{\top}\vec{K}\right)\right)
\end{equation}
where $\vec{y}\in\{-1,1\}^N$ is a vector of concatenated labels and $\text{diag}(.)$ a transformation of a vector into a diagonal matrix. 
Other examples of popular kernel methods include a least squares loss function combined with $\mathcal{L}_2$ regularization, also known as Kernel Ridge Regression and spectral decompositions of the kernel matrix such as kernel PCA \cite{Scholkopf:1998p427}. We refer the interested reader to \cite{shawe2004kernel} for an overview. 

\subsection{Large Scale Kernel Learning}\label{sec:large_scale_kernel_learning}
Evaluating the empirical kernel map in \autoref{eq:kernel_function} for one data point comes at the cost of $N$ evaluations of the kernel function, since the index $j$ (which picks the data points that are used for expanding the {\em kernel map}) runs over all data points in the data set\footnote{Actually the complexity is of $\mathcal{O}(ND)$ where $D$ is the dimensionality of the data, but as this is constant given a data set, we omit this factor here.} both for training and predicting. Computing the full gradient with respect to $\va$ requires $N$ evaluations, too, so the total complexity of computing the gradient of a kernel machine is in the order of $\mathcal{O}(N^2)$. This is the reason why kernel methods became unattractive for large data sets -- as other methods like linear models or neural networks only require training in $\mathcal{O}(N)$ time. 

We categorize attempts to scale up kernel methods into two classes: a) Reducing the number of data points when evaluating the empirical kernel map in \autoref{eq:kernel_function} and b) avoiding to evaluate the empirical kernel map altogether by using an {\em explicit approximation} of the kernel map. There are more sophisticated approaches within each of these categories that can give a quite significant speedup~\cite{Le2013, Rudi2015}. We focus on a comparison between these two types of approximations. We emphasize however that many of these additional improvements also apply to the approach proposed in this manuscript and are likely to improve its convergence and runtime. In the following, we briefly survey research from both categories.

\paragraph{Empirical / implicit kernel maps:} The first approach, reducing the number of data points when evaluating the kernel function, amounts to subsampling data points for computing the empirical kernel map in~\autoref{eq:kernel_function}. 
The data points used to compute the empirical kernel map are sometimes referred to as {\em landmarks}~\cite{Hsieh2014}. 
A prominent line of research in this direction follows what is commonly referred to as the {\em Nystr\"om method}~\cite{Williams2000}. The key idea here is to take a low-rank approximation of the kernel matrix computed on a randomly subsampled part of the data, instead of using the entire matrix. 
Other work in this direction aims at sparsifying the vector of dual coefficients. Another idea similar to our approach is the Naive Online Regularized Risk Minimization Algorithm (NORMA) \cite{Kivinen2004}, the Forgetron \cite{Dekel2008} or other work on online kernel learning. The authors propose ways of speeding up the sum computation by discarding data points from the sum computation in \autoref{eq:kernel_trick}; this is the key difference to the approach proposed here which follows a much simpler randomized scheme.  
Few of the above methods are simple to implement in a parallel or distributed setting. One recent approach is to distribute the data to different workers and solve the kernel problems independently on each worker~\cite{Deisenroth2015}. This implicitly assumes however that the kernel matrix is a block diagonal matrix where the blocks on the diagonal are the kernels on each worker -- all the rest of the kernel matrix is neglected.

\paragraph{Explicit kernel maps:} Recent approaches for large scale kernel learning avoid the computation of the kernel matrix by relying on {\em explicit} forms of the kernel function~\cite{Rahimi2008,Vedaldi2010}. The basic idea is that instead of using a kernel function $k$, (which {\em implicitly} projects the data to kernel feature space $\mathcal{S}$ and computes the inner products in that space in a single step), {\em explicit} kernel functions just perform the first step: mapping to kernel feature space with an approximation of the kernel map $\phi(.)$. This has the advantage of being able to directly control the effective number of features. The model then simply learns a linear combination of these features. Explicit feature maps often express the kernel function as a set of Fourier basis functions. \cite{Dai2014} provides a comprehensive overview of kernel functions and their explicit representations. \cite{Vedaldi2010} gives a more detailed explanation with graphical illustrations for a small set of kernel functions. In the context of large-scale kernel learning this method was popularized by Rahimi and Recht under the name of {\em random kitchen sinks}~\cite{Rahimi2008}. An important parameter choice in these approaches is the number of basis functions. This choice determines the accuracy of the approximation as well as the speed of the computations. 

\paragraph{Which approximation is better?}
Both approaches, {\em implicit kernel maps} and {\em explicit kernel maps}, are similar in that they approximate a mapping to a potentially infinite dimensional space $\mathcal{S}$. The main difference is that for empirical kernel map approaches, the approximation samples data points (and in most cases simply {\em discards} a lot of data points), while in the case of explicit kernel map approximations the approximation samples random Fourier basis functions. 

In practice there are many limitations on how much data can be acquired and processed efficiently. Furthermore, the type of data influences the performance of either approximation: When using the empirical kernel map on extremely sparse data, the empirical kernel function evaluated on a small subset of data points will return 0 in most cases -- while Fourier bases with low frequencies will cover the space of the data much better. 

Which of the two approximations is better in practice is likely to depend on the data. Empirical evidence suggests that the Nystr\"om approximation is better than random kitchen sinks~\cite{Yang2012}. The authors of \cite{Vedaldi2010} perform an extensive comparison of various explicit kernel map approximations and empirical kernel maps, highlighting the advantages in the empirical kernel map approach: Empirical kernel maps have the potential to model some parts of the input distribution better -- but they have to be trained on data. This can be considered a disadvantage. Yet there could be scenarios in which learning the feature representation via empirical kernel maps gives performance gains. We are not aware of a concise comparison of the two approaches in a parallel setting. In our experimental section we provide a direct comparison between the two methods in which we keep the optimization part fixed and concentrate on the type of approximation. 
\section{Doubly stochastic kernel learning}\label{sec:dskl}
This section describes the learning approach to which we refer to as doubly stochastic empirical kernel learning (DSEKL). The key idea is that in each iteration a random sample $\mathcal{I}\subseteq\{1,2,\dots,N\}, |\mathcal{I}|=I$ of data points is chosen for computing the gradient of the dual coefficients $\va$ and another (independent) random sample $\mathcal{J}\subseteq\{1,2,\dots,N\}, |\mathcal{J}|=J$ of data points is chosen for expanding the empirical kernel map $k(.,.)$. Note that this is very similar to Algorithm 1 and 2 in \cite{Dai2014}, except that instead of drawing random basis functions of the kernel function approximation, we sample random data points for expanding the empirical kernel map in~\autoref{eq:kernel_trick}. If one were to compute the entire kernel matrix $\vec{K}\in\R^{N\times N}$, this procedure would correspond to sampling a rectangular submatrix $\vec{K}_{\mathcal{I,J}}\in\R^{I\times J}$. The number of data points $J$ sampled for expanding the empirical kernel map as well as the number of data points $I$ to compute the gradient are important parameters that determine the noise of the gradient of the dual coefficients and the noise of the empirical kernel map, respectively. The pseudocode in algorithm \autoref{alg:dskl} summarizes the procedure, which alternates two steps: 1) sample a random submatrix of the kernel matrix and 2) take a gradient step along the direction of $\frac{\partial\mathcal{E}(\vec{x}_i)}{\partial \va_{j}},~\forall i\in \mathcal{I},~j\in\mathcal{J} $, the gradient of $\mathcal{E}$ w.r.t. $\va$ at indices $\mathcal{J}$ evaluated at data points $\mathcal{I}$.

\begin{algorithm}
  \begin{algorithmic}
    \caption{Doubly Stochastic Kernel Learning\label{alg:dskl}}
     \REQUIRE $(\Bx_i,y_i),i\in\{1,\dots,N\},\Bx_i\in\R^{D},~y_i\in\{-1,+1\}$, Kernel $k(.,.)$
    \ENSURE Dual coefficients $\va$ 
   \STATE \# Initialize coefficients $\va$, initialize counter $i=0$
   \WHILE{Not Converged}
   \STATE $t\gets t+1$
   \STATE \# Sample indices $\mathcal{I}$ for gradient 
   \STATE $\mathcal{I}\sim\text{unif}(1,N)$
   \STATE \# Sample indices $\mathcal{J}$ for empirical kernel map
   \STATE $\mathcal{J}\sim\text{unif}(1,N)$
   \STATE \# Compute Gradient
   \STATE $\forall j \in \mathcal{J}: \quad g_{j} \gets \sum_{i\in\mathcal{I}}\frac{\partial \mathcal{E}(\vec{x}_i)}{\partial\va_{j}}$  (see e.g. \autoref{eq:svm_gradient})
   \STATE \# Update weight vector 
   \STATE $\forall j\in\mathcal{J}: \alpha_j\gets\alpha_j - 1/t~ g_j$ 
   \ENDWHILE
  \end{algorithmic}
\end{algorithm}
Note that in contrast to other kernel approximations, the memory footprint of this algorithm is rather low: While low-rank approximations need to store the low rank factors, algorithm~\autoref{alg:dskl} only requires us to store the dual coefficients $\va$. We simply set the learning rate parameter to $1/t$ where $t$ is the number of iterations. It is good practice to adjust that parameter according to some more sophisticated schedule. We emphasize the applicability of many standard methods to speed up convergence of stochastic gradient, e.g. through better control of the variance of the gradients.
\section{Experiments}\label{sec:experiments}
This experimental section describes experiments on artificial data and on publicly available real-world data sets. All experiments use a support-vector machine with RBF kernel, for the sake of comparability. We performed experiments on a single machine with serial execution (see algorithm \autoref{alg:dskl}) as well as with a parallel shared-memory variant of our approach (see algorithm \autoref{alg:dskl_svm_distributed}). We compared against the batch SVM implementation available in scikit~learn~\cite{sklearn_api}. We conducted experiments with serial execution for small-scale experiments, while we leveraged the parallel variant for larger scale experiments. 

\paragraph{Hyperparameter optimization} We tuned Hyperparameters with two-fold cross-validation and exhaustive grid search for all models; the reported accuracies were computed on a held out test set of the same size as the training set. We select hyperparameters for batch and SGD algorithms (the regularization parameter and RBF scale) from a logarithmic grid from $10^{-6}-10^6$; The SGD approaches have additional parameters such as the step size (candidates were $10^{-4}-10^4$) and the minibatch size $I$ for computing the gradient. For doubly stochastic kernel learning and for random fourier features, there is the additional hyperparameter $J$, referring to the number of kernel expansion coefficients or random fourier features, respectively.

\paragraph{Comparisons with related methods}
We compared the proposed method with other kernel approximations as well as with batch kernel SVMs. We conducted comparisons with random kitchen sinks (RKS) where the number of basis functions matched the number of expansion coefficients $J$. In order to assess standard large-scale kernel approximations that only use a subset of data points, we also compared with a version in which we first draw one random sample from the data, and then train the algorithm with that subset only. While most of these methods that use just a subset of the data apply more sophisticated schemes for selecting that subset and smarter ways of extrapolating, we focused on the main difference here, which is training on a fixed random subset of the data. 

\paragraph{Data sets}
In order to provide a qualitative comparison between the different kernel map approximations, we performed experiments on small synthetic data sets. We chose the XOR problem described in~\autoref{fig:toy_data} as a benchmark for nonlinear classification. We also performed experiments on a number of standard benchmark real world data sets available on the libsvm homepage\footnote{\url{https://www.csie.ntu.edu.tw/~cjlin/libsvmtools/datasets/binary.html}}. \autoref{tab:small_realdata} lists these data sets. 
\begin{figure}[!ht]
    \centering
           \hfill
        \begin{subfigure}[b]{0.5\textwidth}
         Artificial data was generated for a standard two class nonlinear prediction benchmark, the XOR problem.
        Data for one class (yellow dots) is drawn from a spherical gaussian distribution $\mathcal{N}(0,0.2)$ around $[1,1]^\top$ and $[-1,-1]^\top$, and data points from the other class (red dots) are drawn from the same gaussian distribution centered around $[1,-1]^\top$ and $[-1,1]^\top$. 
         Background colors show the classification hyperplane learned by doubly stochastic SVM learning; Larger circles illustrate support vectors.\\     
        \end{subfigure}
	\hfill
         \begin{subfigure}[b]{0.4\textwidth}
         \centering
        \includegraphics[width=1\columnwidth]{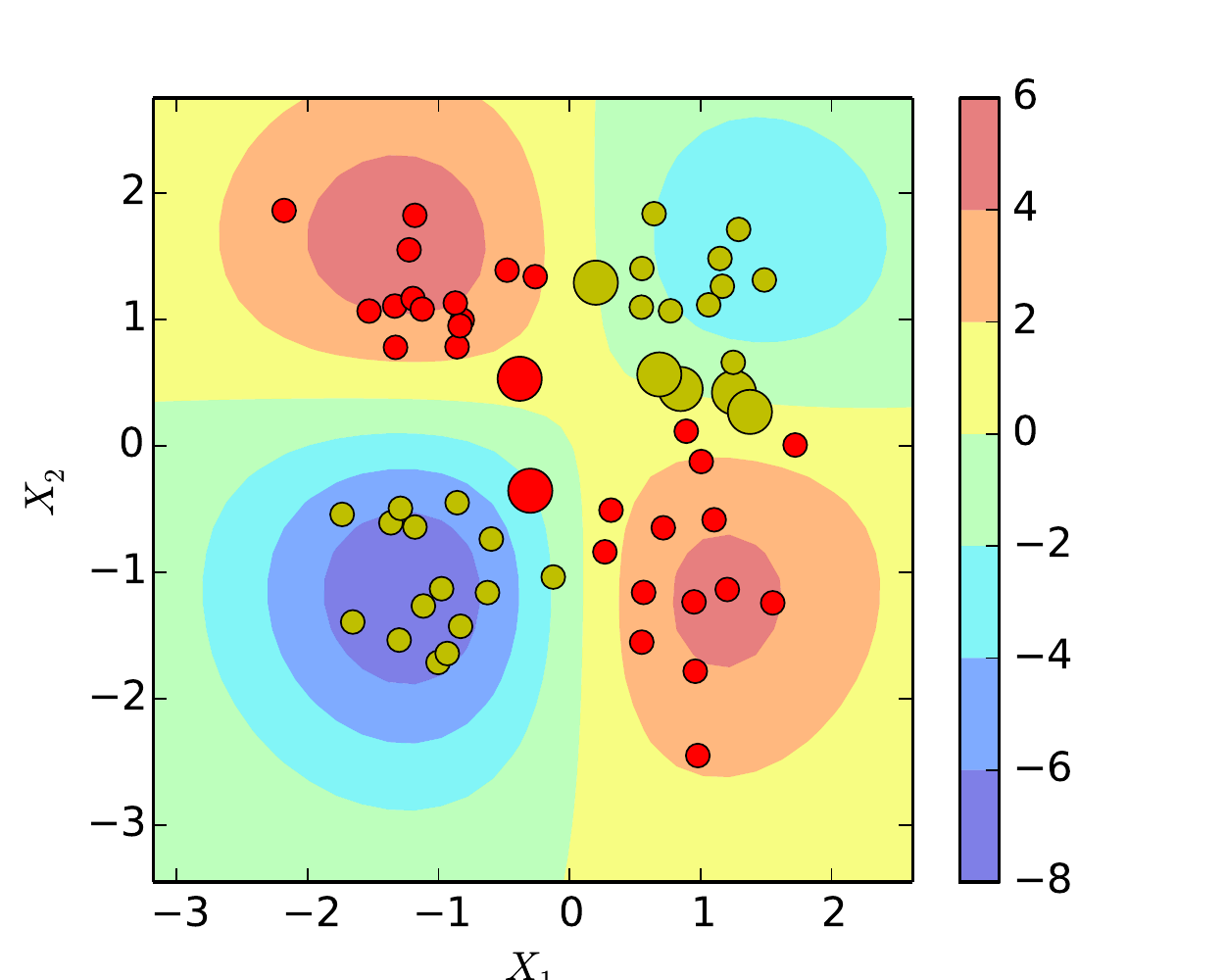}
               \label{fig:toy_data}
        \end{subfigure}

        \caption{\label{fig:toy_data} Synthetic data generation for the XOR problem.}
\end{figure}

\subsection{Serial Execution}\label{sec:single_host}
\begin{figure}[!ht]
   \centering
    \begin{subfigure}[b]{0.235\textwidth}
        \includegraphics[width=\textwidth]{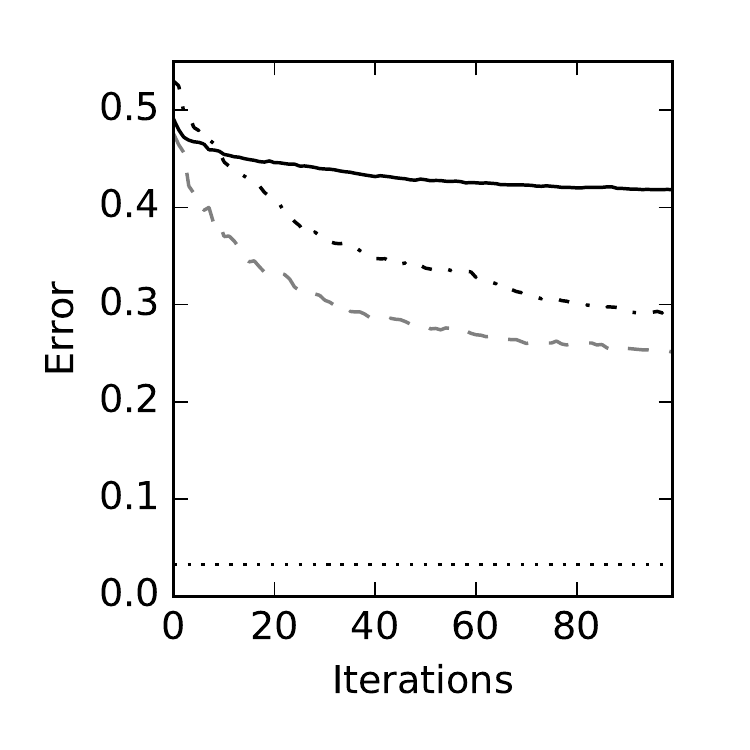}
        \caption{$I=1, J=20$}
        \label{fig:expand_20_pred_1}
    \end{subfigure}
    \hfill
    \begin{subfigure}[b]{0.235\textwidth}
        \includegraphics[width=\textwidth]{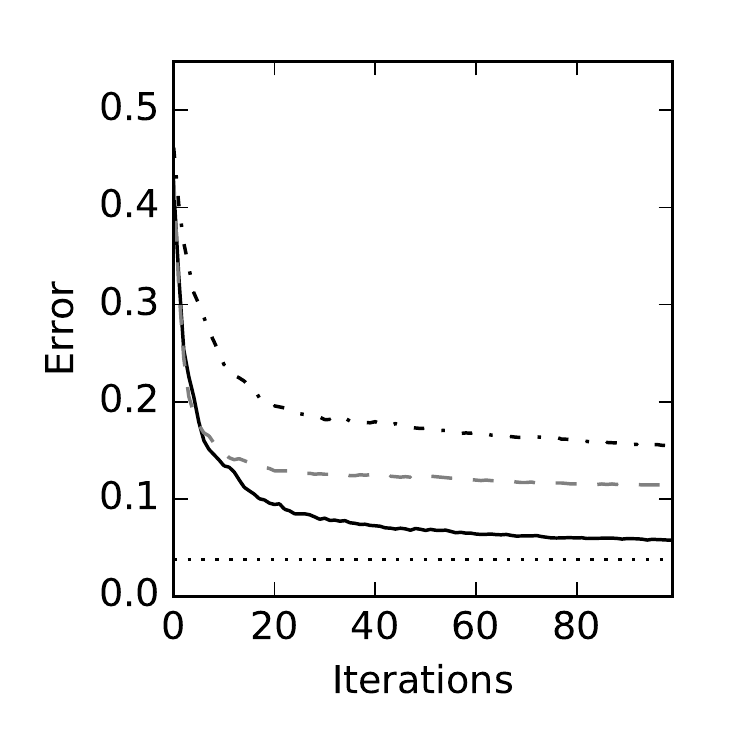}
        \caption{$I=50, J=20$}
        \label{fig:expand_20_pred_50}
    \end{subfigure}
        \begin{subfigure}[b]{0.235\textwidth}
        \includegraphics[width=\textwidth]{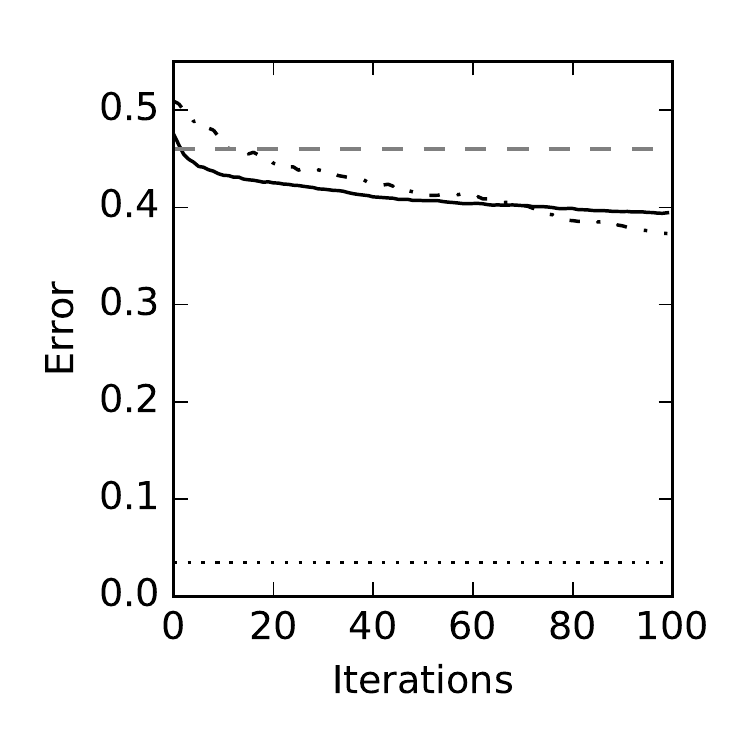}
        \caption{$I=20, J=1$}
        \label{fig:pred_20_expand_1}
    \end{subfigure}
    \hfill
    \begin{subfigure}[b]{0.235\textwidth}
        \includegraphics[width=\textwidth]{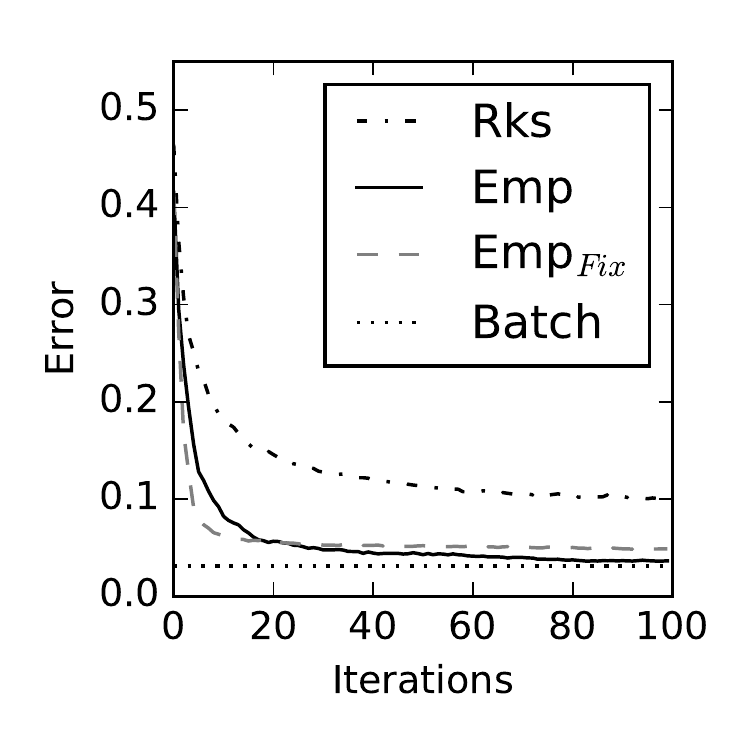}
        \caption{$I=20, J=50$}
        \label{fig:pred_20_expand_50}
    \end{subfigure}
    \caption{Error on test data for XOR problem in \autoref{fig:toy_data} for doubly stochastic kernel learning with the empirical kernel map~(Emp), random kitchen sinks~(RKS), random subsampling~(Emp$_{\text{Fix}}$) and batch SVM; with few expansion samples $J$ and few samples for gradient computations $I$ (\autoref{fig:expand_20_pred_1} and \autoref{fig:pred_20_expand_1}) the explicit kernel map approximations appear to have an advantage; with more samples, doubly stochastic empirical kernel map approximations achieve performances close to batch SVMs (\autoref{fig:expand_20_pred_50} and \autoref{fig:pred_20_expand_50}).}
    \label{fig:toydata_comparisons}
\end{figure}

We ran small-scale experiments with a single threaded implementation of algorithm~\autoref{alg:dskl}. We generated $N=100$ data points according to the XOR problem, see \autoref{fig:toy_data}, and optimized all hyperparameters as described in~\autoref{sec:experiments}.
\autoref{fig:toydata_comparisons} shows comparisons of the proposed method with random kitchen sinks and a fixed random selection of data points, as well as with a batch setting. We plot the error on the test set varying $I$, the number of samples for computing the gradient, in \autoref{fig:expand_20_pred_1} and \autoref{fig:expand_20_pred_50} while keeping all other hyperparameters fixed. \autoref{fig:pred_20_expand_1} and \autoref{fig:pred_20_expand_50} show the error when varying $J$, the number of expansion coefficients. Note that with too few data points for computing the gradient or the expansion, both random kitchen sinks as well as a fixed sample of data points have an advantage over the doubly stochastic approach (\autoref{fig:expand_20_pred_1} and \autoref{fig:pred_20_expand_1}). As the number of data points in the gradient computation and the kernel map expansion increases however, doubly stochastic kernel learning achieves performance comparable to that of batch methods, indicated as dotted line (\autoref{fig:expand_20_pred_50} and \autoref{fig:pred_20_expand_50}).

We performed experiments on a number of standard benchmark real world data sets available on the libsvm homepage\footnote{\url{https://www.csie.ntu.edu.tw/~cjlin/libsvmtools/datasets/binary.html}}. We compared to the batch version using serial execution and small data sets. We discuss experiments on a larger dataset, leveraging our parallel variant in \autoref{sec:distributed}. For all experiments, we sampled $\min(1000,N_{\text{dataset}})$ data points where $N_{\text{dataset}}$ is the number of data points in the respective data set, and took half the data for training and half the data for testing, including hyperparameter optimization on the training set. We ran 10 repetitions of each experiment and show the test set error in \autoref{tab:small_realdata}. In all data sets  investigated, the proposed doubly stochastic empirical kernel learning approach achieved errors comparable to that of a batch SVM. In cases where the batch SVM achieves perfect accuracy and DSEKL still resulted in a few errors, we emphasize that we conduct these comparisons to show that DSEKL has the potential to achieve performance comparable to that of batch methods. Refining the SGD optimization or running more iterations could further improve the performance, yet our main intention is to only provide a proof of concept for the doubly stochastic approach. Also note that the proposed DSEKL approach only uses a fraction of the data in each step. This allows for training on much larger data sets, which we discuss in the next section.
\begin{table}[h!]
\footnotesize
\begin{center}
\begin{tabular}{ ccc } 
Data Set & DSEKL & Batch\\
 \hline
MNIST & 0.00$\pm$0.01  & 0.00$\pm$0.01\\
Diabetes & 0.20$\pm$0.02  & 0.22$\pm$0.02\\
Breast Cancer &  0.03$\pm$0.01 & 0.03$\pm$0.01\\
Mushrooms & 0.03$\pm$0.01 & 0.00$\pm$0.00\\
Sonar & 0.22$\pm$0.07 & 0.26$\pm$0.04\\
Skin/non-skin & 0.03$\pm$0.01 &  0.01$\pm$0.00\\
Madelon & 0.03$\pm$0.01 &  0.00$\pm$0.00\\
 \hline
\end{tabular}
\caption{\footnotesize Test error (mean $\pm$ standard deviation across 10 repetitions) on real world data sets. Doubly stochastic empirical kernel learning (DSEKL) achieves performance comparable to that of a batch kernel SVM. \label{tab:small_realdata}}
\end{center}
\end{table}
 \subsection{Parallel Execution using a Shared-Memory Variant}\label{sec:distributed}
This section describes the experiments performed using a parallel, shared-memory variant of our approach inspired by~\cite{Agarwal2014}. We list the pseudocode in algorithm~\autoref{alg:dskl_svm_distributed}. The difference to algorithm~\autoref{alg:dskl} is that we run multiple workers at the same time, and process multiple sample batches for the empirical kernel map per iteration to parallelize learning. We used sampling without replacement to generate the sample batches for the different workers. 

 \begin{algorithm}[h!]
   \begin{algorithmic}[1]
     \caption{Parallel Shared-Memory Nonlinear Support-Vector Machine\label{alg:dskl_svm_distributed}}
       \REQUIRE sample size $s$, number of workers $K$
       \STATE \# Initialize coefficients $\va$
       \STATE \# Sample indices $\mathcal{I}^{(0)},\dots,\mathcal{I}^{(K)}$ for gradient 
       \STATE \# Sample indices $\mathcal{J}^{(0)},\dots,\mathcal{J}^{(K)}$ for empirical kernel map 
       \STATE $\vec{G} \leftarrow \mathbb{I}$
       \WHILE{Not Converged}
         \FORALL{ $\mathcal{I}^{(0)},\dots,\mathcal{I}^{(K)}$}      
           \FORALL{ $\mathcal{J}^{(0)},\dots,\mathcal{J}^{(K)}$ in parallel on worker $k$} 
             \STATE \# Compute gradients as in Algorithm~\autoref{alg:dskl}
             \STATE $\forall j \in \mathcal{J}^{(k)}: \quad \vec{g}^{(k)}_{j} \gets \sum_{i\in\mathcal{I}^{(k)}}\frac{\partial \mathcal{E}(\vec{x}_i)}{\partial\va_{j}}$
             \STATE \# Aggregate inverse gradients for dampening updates of $\va$
             \STATE $\vec{G}_{ii} \leftarrow \vec{G}_{ii} + \left(g^{(k)}_{ji}\right)^2$ for all $i \in\mathcal{I}^{(k)}$ and $j\in\mathcal{J}^{(k)}$
           \ENDFOR
           \STATE \# Update weight vector 
           \STATE $\va \leftarrow \va - \vec{G}^{-\frac{1}{2}} \sum_k \vec{g}^{(k)}$
         \ENDFOR
       \ENDWHILE
   \end{algorithmic}
 \end{algorithm}

\begin{figure}[h!]
	\begin{subfigure}[b]{0.45\textwidth}
		\centering
		\includegraphics[width=\textwidth]{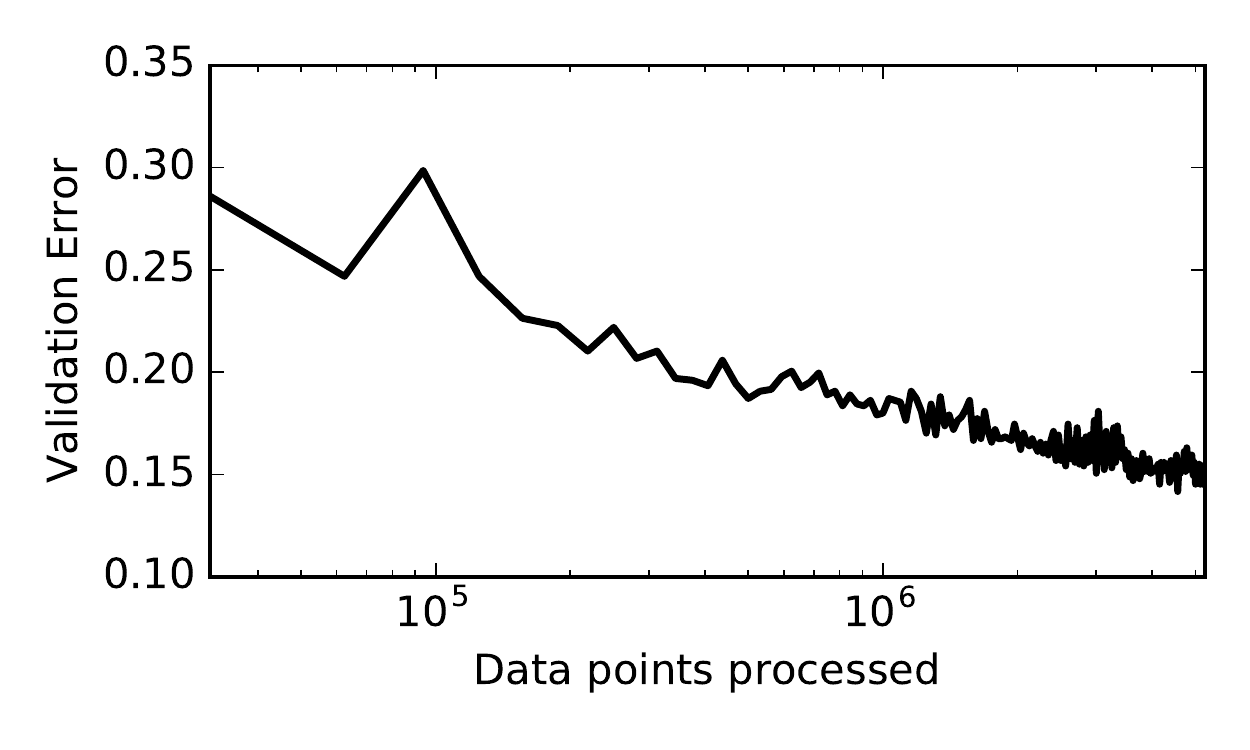}
		\caption{\centering Validation error vs data points processed.}
		\label{fig:covertype_validation}
	\end{subfigure}
	\hfill
	\begin{subfigure}[b]{0.45\textwidth}
		\centering
		\includegraphics[width=\textwidth]{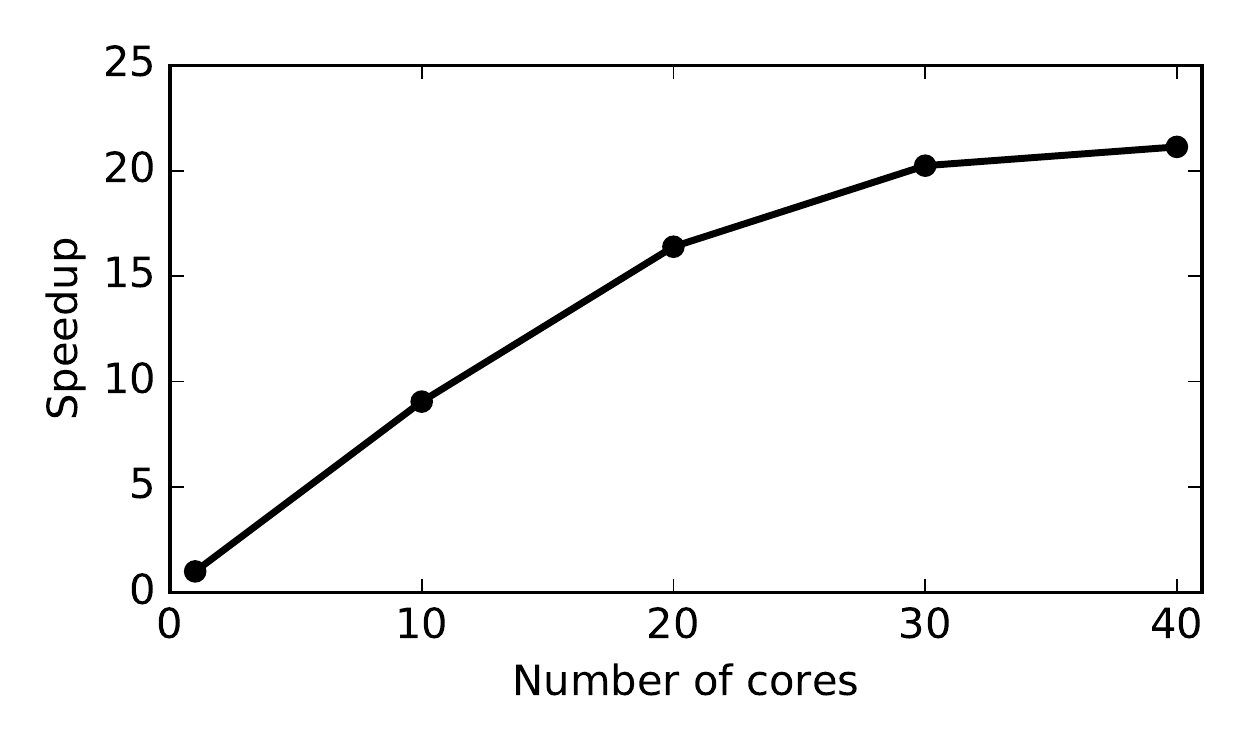}
		\caption{\centering Speedup with increasing number of cores.}
		\label{fig:speedup_58}
	\end{subfigure}
	\caption{\label{fig:large-scale-experiments}Experiments on a larger data set with a parallel implementation of algorithm \autoref{alg:dskl_svm_distributed}. }
\end{figure}

We ran our experiment on the covertype dataset\footnote{\url{https://archive.ics.uci.edu/ml/datasets/Covertype}}, consisting of 581,012 data points with 54 features. We drew samples of $I=10,000$ points for computing the gradient and $J=10,000$ for evaluating the empirical kernel map. For the sake of comparability with the results in~\cite{Dai2014}, we set the regularization parameter $\lambda$ to $1 / N$, and fix the RBF scale to $1.0$. We employ a learning rate of $1/i$, where $i$ is the number of epochs, i.e., passes through the entire data set. We stop the training process if the $\mathcal{L}_{2}$ norm of the weight change over one epoch is less than $1$. We separate the entire data set into three random splits for training, validation during training and evaluation after convergence. For computing the error on the validation set during training, we hold back 1122 random samples. Additionally we hold back a separate random sample of 20,000 data points for the final evaluation after convergence. Figure~\ref{fig:covertype_validation} depicts the validation error after evaluating all $\mathcal{J}$ for one mini batch of $\mathcal{I}$ respectively, for about 3 passes through the whole dataset. After one pass through the data the validation error decreased from 51\% to about 17\%. After 54 epochs the algorithm converged, and the final error rate on the evaluation set was 13.34\%. These results are comparable to~\cite{Dai2014} who report a test error of about 15\% after one iteration. 

Figure \ref{fig:speedup_58} shows the speedup achieved through the usage of multiple cores in our shared-memory variant. 
Our python implementation of algorithm~\autoref{alg:dskl_svm_distributed} runs on a 48 core machine (having 24 physical cores with hyperthreading) with 500 GB main memory. 
We recorded the runtime for processing a single batch $\mathcal{I}^{(k)}$, for which the empirical kernel map is evaluated using all batches $\mathcal{J}^{(k)},~k: 1,\dots,K$ in parallel on twice the full covertype dataset to ensure a full utilization of the machine. We measured speedups against the runtime on a single core and increase the parallelism by ten cores at a time.  We observed a linear speedup until running with 20 cores, where we achieved a speedup of factor 16 compared to the runtime of only one core. After that, the speedup curve flattens out. We attribute this flattening to several overhead factors, such as resource-sharing from hyperthreading after exceeding the number of physical cores, as well as serialization costs caused by python's multithreading. Nevertheless, this experiment shows that our approach amends itself to a simple parallelization scheme, which has the potential for massive speedups.

\section{Conclusion}
When modelling complex functions, the practitioner usually has three main options: Decision-Tree based methods, Neural Networks and Kernel methods. Decision-Tree based methods appear to dominate most Kaggle competitions, and in general give stunning performance on real-world data sets~\cite{Caruana2006}. But when a modelling task goes beyond simple supervised settings, these methods might not be the first choice. Deep neural networks yield large performance improvements on many tasks and are successfully used for unsupervised learning as well -- but they are often difficult to design and train. 
This is where kernel methods offer advantages: Instead of optimizing a network architecture one simply picks an off-the-shelf kernel function for a given type of data, and then one only needs to perform model selection over a handful of kernel parameters in order to tackle both unsupervised and supervised learning in a principled manner.

We have proposed a simple algorithm for scaling up kernel learning that is easy to implement and parallelize. Our results demonstrate that the proposed method achieves competitive performance on standard benchmarks. We hope complementing the existing methods for large scale kernel learning as well as other successful methods such as random forests and neural networks will ultimately help to better understand the strengths of the respective methods, independent of factors such as hardware and optimization procedures. 

Our experiments on artificial data suggest that there are conditions under which the empirical kernel map approach performs better than the explicit kernel map approximation, in agreement with previous results in~\cite{Vedaldi2010}. Yet a direct comparison of our results with results obtained with explicit kernel maps as in \cite{Dai2014} is difficult, due to the differences in the implementations. An important topic of future research will be to investigate when to prefer explicit kernel map approximations as in \cite{Rahimi2008, Dai2014} over the empirical kernel map approaches presented here. In terms of implementation however, applying the doubly stochastic empirical kernel map approach to more complex kernels might appear simpler than implementing a dedicated explicit kernel map approximation for every kernel function. 

Furthermore, we showed that a parallel variant of our algorithm is extremely simple to implement, achieves competitive performance on a large data set, and has the potential for massive speed ups. An interesting direction for the future would be to implement the doubly stochastic approach on graphics cards to leverage their potential for massive parallel computation and use the proposed approach in a streaming/online learning setting, similar to the approaches in \cite{Kivinen2004, Dekel2008} but with a simpler, randomized scheme for reducing the cost of the empirical kernel map computation. Note that it is straightforward to combine the DSEKL approach with truncation schemes as in \cite{Kivinen2004, Dekel2008} during or after convergence for speeding up predictions at test time.

Another interesting direction could be to explore a distributed variant of our algorithm. We found that in its presented form, our approach is not well suited for distributed data processing systems like \textit{Apache Spark}~\cite{Zaharia2012} or \textit{Apache Flink}~\cite{Alexandrov2014} which use a shared-nothing architecture. This is due to the fact that a naive distributed execution of our algorithm would impose a too high amount of communication per iteration for aggregating the gradients over the network, as well as for re-distributing the updated parameter vector. A variant that updates parameters locally on the slaves in the cluster, and only updates the global model from time to time (thereby reducing inter-machine communication) could be worth to look into. 


{\small
  \bibliography{dskl}
  \bibliographystyle{plain}}

\end{document}